\documentclass{ieeeaccess}
\usepackage{cite}
\usepackage{amsmath,amssymb,amsfonts}
\usepackage{algorithmic}
\usepackage{graphicx}
\usepackage{caption}
\usepackage{textcomp}
\usepackage{enumitem}
\usepackage{algorithm2e}
\usepackage{subfigure}
\usepackage{hyperref}
\usepackage{gensymb}
\def\BibTeX{{\rm B\kern-.05em{\sc i\kern-.025em b}\kern-.08em
    T\kern-.1667em\lower.7ex\hbox{E}\kern-.125emX}}
\begin{document}
\history{Date of publication xxxx 00, 0000, date of current version xxxx 00, 0000.}
\doi{10.1109/ACCESS.2017.DOI}

\title{Multiclass Model for Agriculture development using Multivariate Statistical method}

\author{\uppercase{N Deepa}\authorrefmark{1},
	\uppercase{Mohammad Zubair Khan }\authorrefmark{2},
	\uppercase{Prabadevi B}\authorrefmark{1},
	\uppercase{Durai Raj Vincent P M}\authorrefmark{1},
	\uppercase{Praveen Kumar Reddy Maddikunta}\authorrefmark{1},
	\uppercase{Thippa Reddy Gadekallu}\authorrefmark{1}
} 

\address[1]{School of Infromation Technology and Engineering, VIT - Vellore, Tamilnadu, India}
\address[2]{Department of Computer Science, College of Computer Science and Engineering, Taibah University, Madinah, Saudi Arabia}


\markboth
{Deepa \headeretal: Preparation of Papers for IEEE Access}
{Deepa \headeretal: Preparation of Papers for IEEE Access}

\corresp{Corresponding author(s): thippareddy.g@vit.ac.in, mkhanb@taibahu.edu.sa}

\begin{abstract}
	Mahalanobis taguchi system (MTS) is a multi-variate statistical method extensively used for feature selection and binary classification problems. The calculation of orthogonal array and signal-to-noise ratio in MTS makes the algorithm complicated when more number of factors are involved in the classification problem. Also the decision is based on the accuracy of normal and abnormal observations of the dataset. In this paper, a multiclass model using Improved Mahalanobis Taguchi System (IMTS) is proposed based on normal observations and Mahalanobis distance for agriculture development. Twenty-six input factors relevant to crop cultivation have been identified and clustered into six main factors for the development of the model. The multiclass model is developed with the consideration of the relative importance of the factors. An objective function is defined for the classification of three crops, namely paddy, sugarcane and groundnut. The classification results are verified against the results obtained from the agriculture experts working in the field. The proposed classifier provides 100\% accuracy, recall, precision and 0\% error rate when compared with other traditional classifier models.
\end{abstract}


\titlepgskip=-15pt

\maketitle 
\section{Introduction}
\label{Sec1}
Agriculture is a major boon to India, and it is a primary source of income. Though 60\% of the land is cultivable, only 43\% is used for crop production. Farmers in developing countries like India lack proper education and awareness about technical aspects of agriculture land cultivation, crop yield improvement, and soil fertility enhancement. The farmers cultivate their lands based on the previous experiences gained from their ancestors and their own field experiences. But the agriculture land quality parameters have been changing due to the drastic changes in the weather conditions. Also, the fertility of the soil is degraded due to the scarcity of water and rainfall\cite{gadekallu2020novel}.

Due to a lack of awareness on crop cultivation and yield, the farmers who were toiled during the entire cultivation period are paid less because of the mediators (agents for bargaining). If adequate training or assistance for the farmers on crop cultivation, pricing of yield and selling of crops are provided, the hard work laid by the farmers would not go in vain. If not, crop production is reduced, which in turn affects the economy of the country. Therefore if adequate support is provided from the government on these skills apart from the traditional way of doing agriculture, the economy of the country will significantly improve\cite{maddikunta2020unmanned}.

Many countries understand the value of agriculture and have started to shift their focus towards agriculture\cite{vangala2020smart}. They have started to develop enormous innovations in almost all aspect of agriculture like land suitability analysis, soil health monitoring, fertilizer recommendation, good quality seeds, modern farming techniques, advanced irrigation techniques, natural manure production, crop recommendation system, yield prediction, and market price prediction\cite{bojjanovel}. Hence, the government should take proper initiatives (if not in all the areas as mentioned above) to inculcate the importance of agriculture in the young minds from their schooling. In turn, the full experience of our farmers will be transformed into proper techniques and can be utilized for the better health of the forthcoming generation.

In order to provide better recommendations from land suitability analysis to yield prediction, multiple criteria about each area should be considered \cite{1,2}. For instance, apart from the major factors to be considered such as soil, water, fertilizer, seasonal changes some other factors such as distance from agricultural land to research institute, extension centres, markets, agro centres, roads, and seed processing plans should be taken into account. When these factors are considered for better decisions, we must also choose a better technique for this prediction\cite{maddikunta2020predictive}. Henceforth, multi-criteria decision-making(MCDM) models prove to be the best in making decisions from various avenues \cite{26}.

The main contributions of this work are
{
\begin{enumerate}
\item An improved version of Mahalanobis Taguchi system is proposed in this work for multiclass classification problem, and it considers only normal observations of data and applies Mahalanobis distance for classification.
\item The multiclass IMTS model is built by considering the relative importance of the factors, and Grey correlation method is applied to calculate the weights. An objective function is defined to construct the decision matrix for multiclass classification. 
\item Final ranking score matrix is obtained from the objective function to perform multiclass classification of three agriculture crops, and the results are compared with the results of other classifiers such as Naïve Bayes, Decision Tree, Random Forest, AdaBoost, J48, SVM and PART.
\end{enumerate}
}

\section{Related work}
\label{Sec2}
MCDM approaches consider the relative importance of criteria for taking appropriate decisions. Relative importance (the weight) of these criteria plays a significant role in acquiring accurate decisions, and there are many weight calculation methods used with MCDM approaches. A decision model was developed for agriculture development in which Analytical hierarchy process(AHP), Rank sum method, criteria importance through inter-criteria correlation(CRITIC) and Standard Deviation (SD) methods were used for calculating the integrated weights of the criteria. In the development of the decision model, AHP and Rank sum methods are subjective weight assignment methods that calculate weights based on expert input. Further, CRITIC and SD methods are objective weight calculation methods that determine weights through mathematical analysis \cite{2}. A model was developed for susceptibility mapping of floods using Geographical Information System and MCDM approaches. In this model, AHP was used for calculating the weights of eight criteria identified for the development of the model \cite{3}. 

Dominance based rough set approach was specifically used for solving decision-making problems and applied for the development of a decision tool for agriculture development \cite{4}. A model was developed to select materials for manufacturing and design of engineering products using Multi-attribute decision making(MADM) approach namely MULTIMOORA and Shannon entropy method was applied for the calculation of relative importance of the parameters identified for the material selection process \cite{5}. An integrated model was developed using the technique for the order of preference by similarity to ideal solution(TOPSIS), Shannon Entropy and Delphi methods for identification of environmental risk in Iran. Shannon Entropy method was used for the calculation of weights of criteria \cite{6}.

A hybrid decision-making model was developed for the selection of materials for the construction dam by integrating step-wise weight assessment ratio analysis (SWARA) and Combinative distance-based assessment (CODAS) methods. SWARA is a subjective weight calculation method used for the calculation of weights of parameters for the development of the model\cite{8,7}. An MCDM model was developed to monitor the time and attendance of employees in companies. CRITIC method was applied to calculate the weights of the criteria and alternatives were ranked using MCDM method, namely Weighted Aggregated Sum Product Assessment (WASPAS) method\cite{8}.

A decision model was built for ranking the journals using TOPSIS method by applying two-weight calculation methods, namely Rough set approach\cite{tripathy2008rough} and  Grey correlation method \cite{9}. Grey Relational Analysis (GRA) is a popular MCDM method used for decision making when multiple conflicting criteria are involved. A ranking model was developed to evaluate the energy consumption in 47 official buildings using MCDM approach, namely GRA. GRA is specifically used to handle the relationships between multiple criteria considered for the problem and uncertain data\cite{10}. Further Grey correlation method was employed for the calculation of weights of factors identified for the development of the multiclass model.

Mahalanobis Taguchi System(MTS) is a multivariate statistical method gaining popularity in decision-making problems. It was introduced by Prof. Mahalanobis, who discovered Mahalanobis Distance(MD) in 1930  to identify the sample from a given set of samples\cite{11}. MTS has been used nowadays to select useful set of variables from the available set of identified variables for decision-making problems\cite{12,13}. A disease classification model was built using MTS, Fuzzy approach and C-Means clustering algorithm. MTS was applied for the selection of attributes from the dynamically selected features of Electrocardiogram(ECG)\cite{14}.

{Some of the metric-learning based methods also use MD to solve the complex decision problems. To avoid the ill-conditioned formulations in hyperspectral images(HIS) distance metric learning is used for dimensionality reduction of the HIS images. A discriminative local metric learning method was developed in \cite{ dong2017dimensionality}, to attain a global metric learning method for dimensionality reduction of HIS. Similarly, deep distance metric learning was proposed in \cite{zhe2019directional} using convolution neural network for image classification. L2-normalization with cosine similarity was employed to improve the performance of the model.}

An optimized binary classification was developed using a modified MTS(MMTS) method. The MMTS showed better results compared with the results obtained from Support Vector Machine(SVM), Probabilistic MTS, Naive Bayes, Hidden Naive Bayes, Kernel Boundary Alignment, Adaptive Conformal Transformation and Synthetic Minority Oversampling Technique methods\cite{15,jindal2020internet,patel2020review}. A novel method was developed for the identification of conditions of roads using MTS, where it was applied to classify the quality of roads in cities\cite{16}. A novel decision model was built using MTS for the classification of gait patterns for the patients treated for ligament reconstruction. Here MTS was used for both feature selection and classification purposes\cite{17}. 

An evaluation model was developed for ranking the dangerous chemicals using Mahalanobis Taguchi System method. A multivariate analysis was done on the dataset and correlation among the criteria were considered for ranking the given set of alternatives\cite{18}. An evaluation model was built to rank the performance of energy security\cite{azab2014mining,alazab2013malicious} in China using 14 factors that are relevant to the problem. Mahalanobis Taguchi Gram Schmidt was applied for the calculation of weights of identified factors, and TOPSIS was used to rank the given set of alternatives\cite{19}. MTS has been proved successful in binary classification. But it has been improved further for the classification of multiclass data also. {Several models have been developed for the detection of faults in various devices and equipment in mechanical domain \cite{ wu2017descriptor}, \cite{ wu2020incipient}.}

An adaptive multiclass MTS model was developed for the identification of faults in bearings \cite{20}.{
Furthermore, various emerging models were used for classification from a larger image dataset. A linear classification system was developed in \cite{ tseng2020interpretable} using maximum a posterior on face recognition dataset. The data were compressed using dimensionality reduction techniques to enhance the performance of the model. The model achieved better results in low computational complexity (reduced training and testing time) and better accuracy of 97.61\% when compared to existing conventional methods. Similarly, A feature learning model was developed in \cite{ luo2018feature }  for a hyperspectral image (HSI) containing a vast number of spatial-spectral information. The feature learning model using spatial-spectral information, hypergraph learning and discriminant analysis improves the performances of the classification to a greater extent when compared to conventional methods. Also, a dimensionality reduction technique with discriminant learning for enhancing the classification accuracy of HIS was developed in \cite{ luo2020dimensionality}. The model outperforms the other dimensionality reduction techniques by exhibiting the complicated intrinsic relationships of HIS.}

A multi-objective firework algorithm was proposed for automatic clustering and classification which contains dynamic searching feature, remodelled objective function for clustering, modified mutual data and automatic clustering capability \cite{li2020learning}. A dividing based objective evolutionary algorithm was proposed for feature selection on huge dataset. Two wrapper and filter was designed for obtaining high accuracy and achieving low computation cost. In order to obtain fast convergence, two recombination techniques were presented. A triangular decision making was also proposed using manhattan distance metric for providing assistance to the users knowledge \cite{li2019dividing}. A feature based data exchange facility is proposed in cloud based design and manufacturing domain \cite{wu2015service}.

In this paper, an Improved Mahalanobis Taguchi method was applied to develop a multiclass model for the classification of three agriculture crops. MTS was basically used for classification purpose by considering normal and abnormal observations relevant to the problem. In this multiclass model, the usage of abnormal observations is not required for classification, and the proposed Improved MTS method is simple and requires a limited number of calculations to perform classification. Also, the use of Mahalanobis distance value for each crop improves the distinguishability among them.

The rest of the paper is organized as follows:  Section \ref{Sec3} discusses the proposed model, identified factors, study area and dataset used in the paper. Section \ref{sec4} explains the results of Grey correlation method, objective function and Improved MTS. The paper is concluded in \ref{sec5}.

\section{Material and methods}
\label{Sec3}
\subsection{Proposed Model}
\begin{figure}[h!]
 	\centering
 	\includegraphics[width=\linewidth]{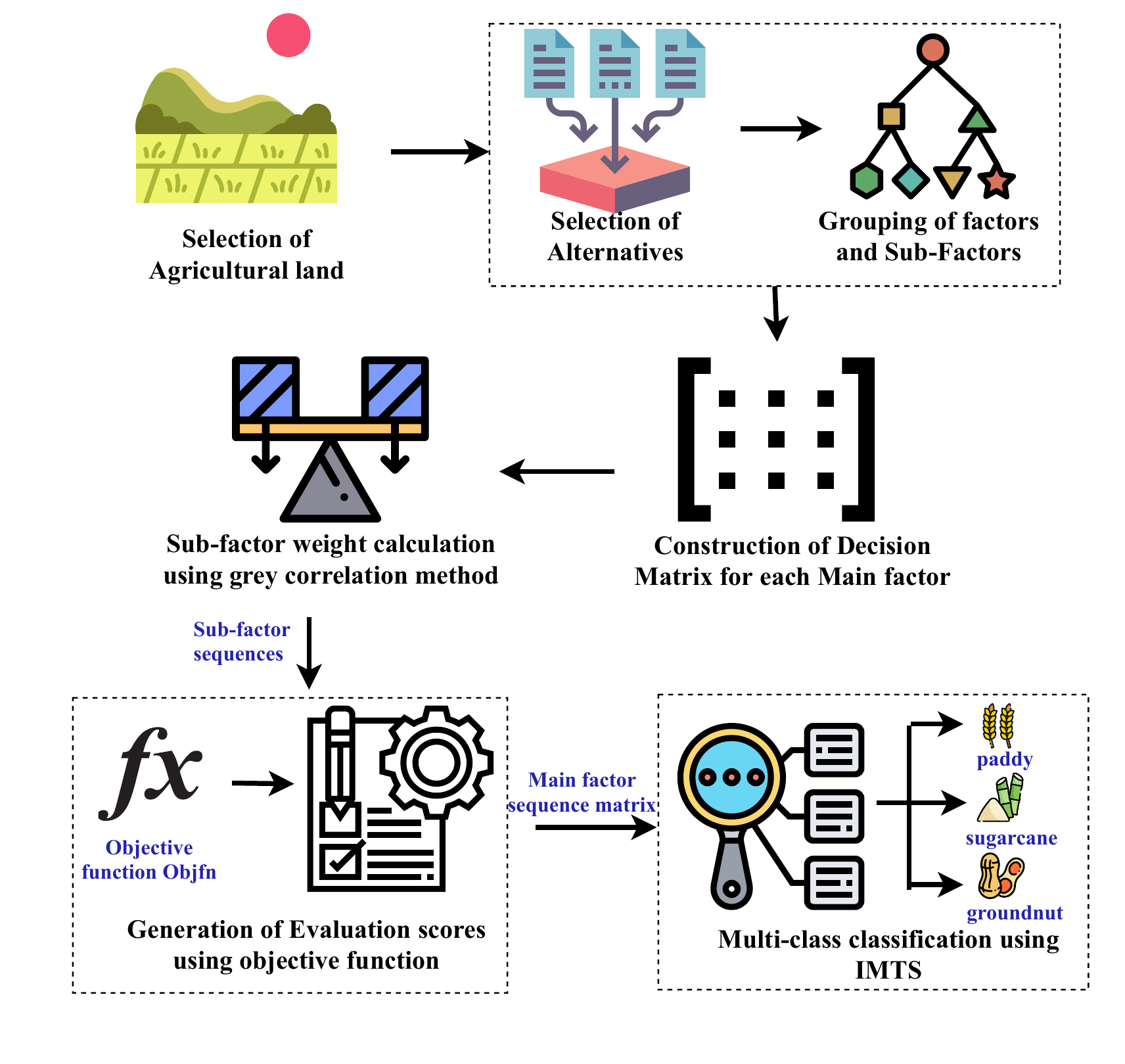}
 	\caption{{Schematic diagram of the proposed multiclass IMTS model.}}
 	\label{fig:fig1}
 \end{figure}

 The architecture of the proposed multiclass IMTS model is shown in Figure \ref{fig:fig1}.
 
The Proposed multiclass model is segregated into six stages:
\begin{enumerate}
	\item Selection of experimental land.
	\item Identification of relevant factors and crops pertaining to the given problem.
	\item Construction of decision matrix for each main factor for the given crops.
	\item Computation of weights of sub-factors under each main factor using Grey correlation method.
	\item Generation of evaluation scores by applying objective function which transforms sub-factor sequences to main factor sequence matrix.
	\item Classification of main factor sequence matrix for three crops using Improved MTS.
\end{enumerate}

{A multiclass model for decision making on crop selection for the given agricultural site with influential parameters is proposed to assist the farmer in gaining the utmost profit by maximizing their yield. Agriculture land is selected where paddy, groundnut and sugarcane crops are cultivated as major crops. Though this classification model can provide better decisions on any crop selection, here three crops viz., paddy, groundnut, and sugarcane are chosen, for which these twenty-six input factors are obtained from the identified experimental land and through the survey.  As there are many factors considered, they are clustered into six main factors viz., soil(mf1), water(mf2), season(mf3), fertilizer-input(mf4), support(mf5) and amenities(mf6). A decision matrix is constructed for each main factor where all the sub-factor values are included. }

{
	As weight calculation plays an important role in decision making, weights are computed for the sub-factors in each main factor using Grey correlation method. The collected agriculture site dataset consists of different values of measurements, and therefore, data normalization is performed. An objective function is defined to generate the evaluation scores, which are normalized values of the raw data collected. Also, the objective function transforms the sub-factor values into main factor values using the weights of the sub-factors and the sub-factor decision matrix. The evaluation scores of the three crops, namely paddy, groundnut and sugarcane, are applied to the Improved Mahalanobis Taguchi method for classification. The proposed method determines the suitability of a crop that can be cultivated in the given agriculture site and the performance of the model is validated by the classification results carried out by the agriculture field experts for the same dataset.
}

\subsection{Identification of factors and sub-factors}
\begin{figure}[h!]
	\centering
	\includegraphics[width=\linewidth]{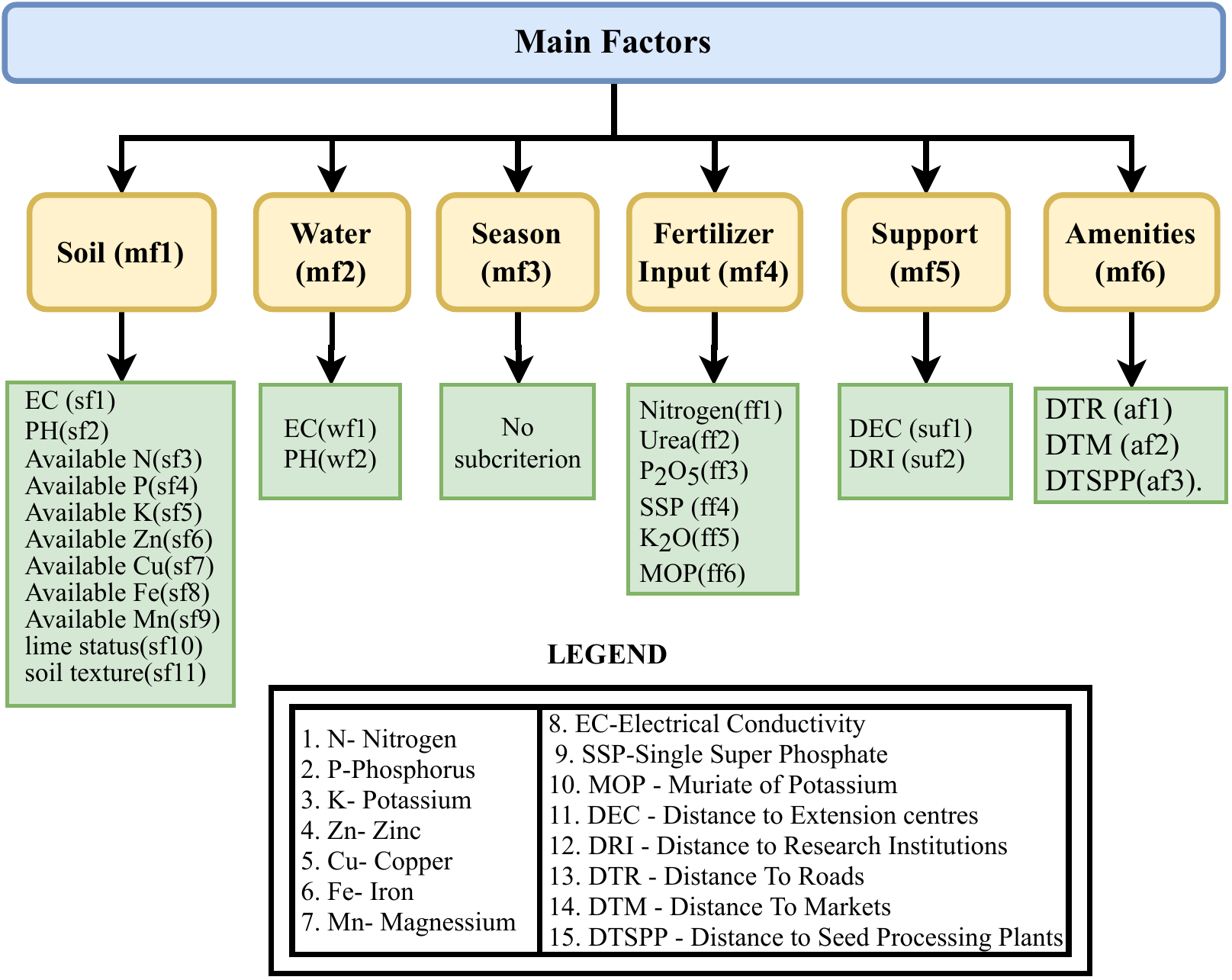}
	\caption{{Main factors and corresponding sub-factors identified for the multiclass model.}}
	\label{fig:fig2}
\end{figure}

Based on the agriculture field experts' opinion and from analysis of the literature survey done, 26 factors were identified for the development of the proposed IMTS model. Further 26 factors were clustered into six main factors each of which have its own sub-factors viz., soil(11 sub-factors), water(2 sub-factors), season(no sub-factor), fertilizer-input(6 sub-factors), support(2 sub-factors) and amenities(3 sub-factors) as shown in Figure \ref{fig:fig2}.

\subsection{Study Area and Data sets}
The field of study was Tiruvannamalai district in the state of Tamil Nadu, India. As mentioned above, the three crops namely paddy, groundnut, and sugarcane, are chosen for experimental purposes and considered as major economic crops in the geographical area with latitude, $12^\circ$15$^{'}N$ and the longitude, $79^\circ$07$^{'}E$. The agricultural sites from the various village panchayats of Tiruvannamalai block namely Melkachirapattu, Thalayampallam, Andampallam, Allikondapattu, Devanur and Perumanam were chosen randomly for collecting the dataset for the study. The chosen main factors namely soil(mf1), water(mf2), season(mf3), fertilizer-input(mf4), support(mf5) and amenities(mf6) associated data for chosen three crops were collected for the development of the multiclass model. Out of 15 sites, three sites pertain to paddy crop, three sites related to sugarcane crop and remaining 3 sites to groundnut crop. Thus decision matrix comprises of sub-factor values under each main factor for each crop is constructed from the raw data for the development of the model.
\section{Results and Discussions}
\label{sec4}
\subsection{Calculation of weights of sub-factors using Grey correlation method}
Grey correlation method is applied to each sub-factor decision matrix which consists of raw data for calculation of relative weights. The first step in Grey correlation method is the generation of comparability sequences. As the raw data consists of a different range of values, it is advisable to normalize the values to the same measurement values. Comparability sequence consists of normalized values of the original decision matrix, which is calculated using the formula given as follows: 
\begin{equation}Y_{i j}=\left(X_{i j}-\min \left(X_{i j}\right)\right) /\left(\max \left(X_{i j}\right)-\min \left(X_{i j}\right)\right)\end{equation}
Where $X_{ij}$ is the sub-factor matrix, $i=1,2,3,…m$, $j=1,2,3,…n$, and $m$ is number of alternatives (agriculture site dataset) and $n$ is number of sub-factors in given main factor. 
The comparability sequence matrix for sub-factors under soil main factor is shown in Table \ref{tab:table1}.
\begin{table*}[h!]
	\centering
	\caption{Comparability Sequence matrix of sub-factors under soil main factor.}
	\label{tab:table1}
	\begin{tabular}{|c|c|c|c|c|c|c|c|c|c|c|}
		\hline
		\textbf{Sf1} & \textbf{Sf2} & \textbf{Sf3} & \textbf{Sf4} & \textbf{Sf5} & \textbf{Sf6} & \textbf{Sf7} & \textbf{Sf8} & \textbf{Sf9} & \textbf{Sf10} & \textbf{Sf11} \\ \hline
		0.00 & 1.00 & 0.0000 & 0.7778 & 0.0000 & 0.0000 & 0.0000 & 0.0000 & 0.7101 & 0.00 & 1.00 \\ \hline
		0.5  & 0.50 & 1.0000 & 1.0000 & 0.7368 & 0.8182 & 0.0606 & 0.2837 & 1.0000 & 1.00 & 0.00 \\ \hline
		0.00 & 1.00 & 1.0000 & 0.0000 & 0.9474 & 1.0000 & 0.0000 & 0.4382 & 1.0000 & 0.00 & 1.00 \\ \hline
		1.00 & 0.00 & 0.2609 & 0.0556 & 1.0000 & 0.0000 & 1.0000 & 1.0000 & 0.0000 & 1.00 & 0.00 \\ \hline
		0.00 & 1.00 & 0.2609 & 0.0556 & 1.0000 & 0.0000 & 1.0000 & 1.0000 & 0.0000 & 0.00 & 1.00 \\ \hline
	\end{tabular}
\end{table*}
The reference sequence is defined as  
\begin{equation}Y_{0 j}=[11111 \ldots 1]\end{equation}
which is the ideal solution for the given alternatives. The next step is the computation of grey relational degree, which gives the distance between the ideal solution and the comparability sequence. Grey relational degree is calculated using the formula given as follows:
\begin{equation}\delta_{i j}=\left|Y_{0 j}-Y_{i j}\right|\end{equation}
Grey relational degree values for sub-factors of soil main factor are presented in Table \ref{tab:table2}. 

\begin{table*}[h!]
	\centering
	\caption{Grey relational degree values of sub-factors under soil main factor.}
	\label{tab:table2}
	\begin{tabular}{|c|c|c|c|c|c|c|c|c|c|c|}
		\hline
		\textbf{Sf1} & \textbf{Sf2} & \textbf{Sf3} & \textbf{Sf4} & \textbf{Sf5} & \textbf{Sf6} & \textbf{Sf7} & \textbf{Sf8} & \textbf{Sf9} & \textbf{Sf10} & \textbf{Sf11} \\ \hline
		0.00 & 1.00 & 0.00   & 0.7778 & 0.0000 & 0.0000 & 0.0000 & 0.0000 & 0.7101 & 0.00 & 1.00 \\ \hline
		0.50 & 0.50 & 1.00   & 1.0000 & 0.7368 & 0.8182 & 0.0606 & 0.2837 & 1.0000 & 1.00 & 0.00 \\ \hline
		0.00 & 1.00 & 1.00   & 0.0000 & 0.9474 & 1.0000 & 0.0000 & 0.4382 & 1.0000 & 0.00 & 1.00 \\ \hline
		1.00 & 0.00 & 0.2609 & 0.0556 & 1.0000 & 0.0000 & 1.0000 & 1.0000 & 0.0000 & 1.00 & 0.00 \\ \hline
		0.00 & 1.00 & 0.2609 & 0.0556 & 1.0000 & 0.0000 & 1.0000 & 1.0000 & 0.0000 & 0.00 & 1.00 \\ \hline
	\end{tabular}
\end{table*}
Next Grey coefficient values are calculated using the equation given as follows:
\begin{equation}C_{i j}=\left(\delta_{\min }+\left(t h^{*} \delta_{\max }\right)\right) /\left(\delta_{i j}+\left(t h^{*} \delta_{\max }\right)\right)\end{equation}
Where $\delta_{\max }=\max \left(\delta_{i j}\right)$  and $\delta_{\min }=\min \left(\delta_{i j}\right)$  and th is threshold value which is a unique coefficient number which spans between 0 and 1. The threshold value is defined as 0.5 for most of the problems in MCDM \cite{21}. Thus Grey relational coefficient values calculated for each alternative for soil main factor are shown in Table \ref{tab:table3}.
\begin{table*}[h!]
	\centering
	\caption{Grey coefficient values of sub-factors under soil main factor.}
	\label{tab:table3}
	\begin{tabular}{|c|c|c|c|c|c|c|c|c|c|c|}
		\hline
		\textbf{Sf1} & \textbf{Sf2} & \textbf{Sf3} & \textbf{Sf4} & \textbf{Sf5} & \textbf{Sf6} & \textbf{Sf7} & \textbf{Sf8} & \textbf{Sf9} & \textbf{Sf10} & \textbf{Sf11} \\ \hline
		1.00 & 0.00 & 1.0000 & 0.2222 & 1.0000 & 1.0000 & 1.0000 & 1.0000 & 0.2899 & 1.00 & 0.00 \\ \hline
		0.50 & 0.50 & 0.0000 & 0.0000 & 0.2632 & 0.1818 & 0.9394 & 0.7163 & 0.0000 & 0.00 & 1.00 \\ \hline
		1.00 & 0.00 & 0.0000 & 1.0000 & 0.0526 & 0.0000 & 1.0000 & 0.5618 & 0.0000 & 1.00 & 0.00 \\ \hline
		0.00 & 1.00 & 0.7391 & 0.9444 & 0.0000 & 1.0000 & 0.0000 & 0.0000 & 1.0000 & 0.00 & 1.00 \\ \hline
		1.00 & 0.00 & 0.7391 & 0.9444 & 0.0000 & 1.0000 & 0.0000 & 0.0000 & 1.0000 & 1.00 & 0.00 \\ \hline
	\end{tabular}
\end{table*}
Correlation degree values are calculated for the alternatives of sub-factors under each main factor for the identified crops using the formula given as follows:
\begin{equation}C_{j}=\frac{1}{n} \sum_{i=1}^{n} C_{i j}\end{equation}
The correlation degree values obtained for each alternative for soil main factor are shown in Table \ref{tab:table4}.
\begin{table*}[h!]
	\centering
	\caption{Correlation degree values of sub-factors under soil main factor.}
	\label{tab:table4}
	\begin{tabular}{|c|c|c|c|c|c|c|c|c|c|c|}
		\hline
		\textbf{Sf1} & \textbf{Sf2} & \textbf{Sf3} & \textbf{Sf4} & \textbf{Sf5} & \textbf{Sf6} & \textbf{Sf7} & \textbf{Sf8} & \textbf{Sf9} & \textbf{Sf10} & \textbf{Sf11} \\ \hline
		0.3333 & 1.0000 & 0.3333 & 0.6923 & 0.3333 & 0.3333 & 0.3333 & 0.3333 & 0.6330 & 0.3333 & 1.0000 \\ \hline
		0.5000 & 0.5000 & 1.0000 & 1.0000 & 0.6552 & 0.7333 & 0.3474 & 0.4111 & 1.0000 & 1.0000 & 0.3333 \\ \hline
		0.3333 & 1.0000 & 1.0000 & 0.3333 & 0.9048 & 1.0000 & 0.3333 & 0.4709 & 1.0000 & 0.3333 & 1.0000 \\ \hline
		1.0000 & 0.3333 & 0.4035 & 0.3462 & 1.0000 & 0.3333 & 1.0000 & 1.0000 & 0.3333 & 1.0000 & 0.3333 \\ \hline
		0.3333 & 1.0000 & 0.4035 & 0.3462 & 1.0000 & 0.3333 & 1.0000 & 1.0000 & 0.3333 & 0.3333 & 1.0000 \\ \hline
	\end{tabular}
\end{table*}
The relative weights of sub-factors are obtained by normalizing the correlation degree values using the formula
\begin{equation}w_{j}=\frac{C_{j}}{\sum_{j=1}^{n} C_{j}}\end{equation}
Thus the relative weights of sub-factors under each main factor are tabulated and shown in Table \ref{tab:table5}.
\begin{table*}[h!]
	\centering
	\caption{{Weights of sub-factors under each main factor.}}
	\label{tab:table5}
	\begin{tabular}{|c|c|cccccccc}
		\hline
		\textbf{mf1} &
		\textbf{weights} &
		\multicolumn{1}{c|}{\textbf{mf2}} &
		\multicolumn{1}{c|}{\textbf{weights}} &
		\multicolumn{1}{c|}{\textbf{mf3}} &
		\multicolumn{1}{c|}{\textbf{weights}} &
		\multicolumn{1}{c|}{\textbf{mf4}} &
		\multicolumn{1}{c|}{\textbf{weights}} &
		\multicolumn{1}{c|}{\textbf{mf5}} &
		\multicolumn{1}{c|}{\textbf{weights}} \\ \hline
		{\textbf{sf1}} &
		0.0714 &
		\multicolumn{1}{c|}{\textbf{wf1}} &
		\multicolumn{1}{c|}{0.4973} &
		\multicolumn{1}{c|}{\textbf{ff1}} &
		\multicolumn{1}{c|}{0.2022} &
		\multicolumn{1}{c|}{\textbf{suf1}} &
		\multicolumn{1}{c|}{0.5345} &
		\multicolumn{1}{c|}{{\textbf{af1}}} &
		\multicolumn{1}{c|}{0.3483} \\ \hline
		{\textbf{sf2}} &
		0.1095 &
		\multicolumn{1}{c|}{{\textbf{wf2}}} &
		\multicolumn{1}{c|}{0.5027} &
		\multicolumn{1}{c|}{\textbf{ff2}} &
		\multicolumn{1}{c|}{0.1816} &
		\multicolumn{1}{c|}{{\textbf{suf2}}} &
		\multicolumn{1}{c|}{0.4655} &
		\multicolumn{1}{c|}{{\textbf{af2}}} &
		\multicolumn{1}{c|}{0.3033} \\ \hline
		{\textbf{sf3}} &
		0.0897 &
		&
		\multicolumn{1}{c|}{} &
		\multicolumn{1}{c|}{\textbf{ff3}} &
		\multicolumn{1}{c|}{0.1847} &
		&
		\multicolumn{1}{c|}{} &
		\multicolumn{1}{c|}{{\textbf{af3}}} &
		\multicolumn{1}{c|}{0.3483} \\ \cline{1-2} \cline{5-6} \cline{9-10} 
		{\textbf{sf4}} &
		0.0776 &
		&
		\multicolumn{1}{c|}{} &
		\multicolumn{1}{c|}{\textbf{ff4}} &
		\multicolumn{1}{c|}{0.1846} &
		&
		&
		&
		\\ \cline{1-2} \cline{5-6}
		{\textbf{sf5}} &
		0.1112 &
		&
		\multicolumn{1}{c|}{} &
		\multicolumn{1}{c|}{\textbf{ff5}} &
		\multicolumn{1}{c|}{0.1156} &
		&
		&
		&
		\\ \cline{1-2} \cline{5-6}
		{\textbf{sf6}} &
		0.0781 &
		&
		\multicolumn{1}{c|}{} &
		\multicolumn{1}{c|}{\textbf{ff6}} &
		\multicolumn{1}{c|}{0.1313} &
		&
		&
		&
		\\ \cline{1-2} \cline{5-6}
		{\textbf{sf7}} &
		0.0861 &
		&
		&
		&
		&
		&
		&
		&
		\\ \cline{1-2}
		{\textbf{sf8}} &
		0.0918 &
		&
		&
		&
		&
		&
		&
		&
		\\ \cline{1-2}
		{\textbf{sf9}} &
		0.0942 &
		&
		&
		&
		&
		&
		&
		&
		\\ \cline{1-2}
		{\textbf{sf10}} &
		0.0857 &
		&
		&
		&
		&
		&
		&
		&
		\\ \cline{1-2}
		{\textbf{sf11}} &
		0.1047 &
		&
		&
		&
		&
		&
		&
		&
		\\ \cline{1-2}
	\end{tabular}
\end{table*}

\subsection{Construction of objective function}
The input sub-factor matrix under each main factor and their relative weights are applied to objective function in order to rank the given set of alternatives with respect to their main factors. In other words, sub-factor sequence values can be combined to form main factor matrix values in the form of ranking scores assigned for each alternative using the objective function. Thus an objective function is defined using the sub-factor matrix values and relative weights of the sub-factors obtained using the Grey correlation method.
\begin{equation}\textit {Objfn}_{i}=\sum_{j} D_{i j} w_{j}\end{equation}

Where $D_{ij}$ is the sub-factor decision matrix obtained from the experimental dataset for the identified crops and wj is the weights of sub-factors under each main factor. $i=1,2,3,…m$, $j=1,2,3,…n$. $m$ is the number of alternatives, $n$ is the number of factors.
Each sub-factor matrix, along with its weights, is applied to the objective function, and final ranking scores are obtained. As the main factor (mf3) input does not have sub-factor and hence the corresponding alternative values are normalized and included in the ranking score decision matrix. Thus obtained main factor matrix for the identified 3 crops, namely paddy, sugarcane and groundnut is shown in Table \ref{tab:table6}.
\begin{table*}[h!]
	\centering
	\caption{Final ranking scores of alternatives for 3 crops.}
	\label{tab:table6}
	\begin{tabular}{|c|c|c|c|c|c|c|}
		\hline
		\textbf{mf1} & \textbf{mf2} & \textbf{mf3} & \textbf{mf4} & \textbf{mf5} & \textbf{mf6} & \textbf{decision class} \\ \hline
		0.5901 & 0.716  & 0.9    & 1.4668 & 0.5725 & 0.8268 & paddy     \\ \hline
		1.1521 & 0.6835 & 0.9021 & 1.2722 & 0.1352 & 0.3435 & paddy     \\ \hline
		1.1704 & 0.5707 & 0.9021 & 1.2654 & 0.1418 & 0.225  & paddy     \\ \hline
		0.7887 & 0.4867 & 0.9    & 1.5224 & 0.1176 & 0.3238 & paddy     \\ \hline
		0.7887 & 0.4426 & 0.9021 & 1.5224 & 0.5725 & 0.8268 & paddy     \\ \hline
		0.34   & 0.6916 & 0.2919 & 1.6126 & 0.9583 & 2.0364 & sugarcane \\ \hline
		0.868  & 0.3832 & 0.2919 & 1.6467 & 0.6667 & 1.3818 & sugarcane \\ \hline
		0.616  & 0.7105 & 0.3    & 1.1793 & 0.0417 & 0.1455 & sugarcane \\ \hline
		0.568  & 0.4295 & 0.2919 & 1.1022 & 0.5417 & 1.1636 & sugarcane \\ \hline
		0.816  & 0.7368 & 0.2919 & 2.6178 & 0.9583 & 2.0364 & sugarcane \\ \hline
		0.4852 & 0.6156 & 1.259  & 0.4769 & 1.0752 & 1.8172 & groundnut \\ \hline
		0.5074 & 0.2208 & 0.7834 & 0.3527 & 0.1466 & 0.2796 & groundnut \\ \hline
		0.387  & 0.5481 & 1.259  & 0.4967 & 0.4887 & 0.9086 & groundnut \\ \hline
		0.587  & 0.6156 & 0.7834 & 0.4363 & 0.6842 & 1.1882 & groundnut \\ \hline
		0.5981 & 0.2208 & 0.7834 & 0.3527 & 0.3421 & 0.5591 & groundnut \\ \hline
	\end{tabular}
\end{table*}

\subsection{Classification using Improved Mahalanobis Taguchi System }
Mahalanobis Taguchi system is a statistical method used for classification purpose. It considers normal and abnormal observations relevant to the problem. In this problem, normal observations are the agriculture site dataset suitable for crop cultivation, and abnormal observations are the agriculture sites which are not suitable for cultivation. In this Improved MTS method, the usage of abnormal observations is not required for classification. And the proposed Improved MTS method is simple and requires a limited number of calculations to perform classification. It uses mahalanobis distance value for each crop to distinguish among them.

The obtained final ranking scores of 15 alternatives for 3 crops, namely paddy, sugarcane and groundnut are applied to Improved MTS algorithm for classification of 3 crops. The steps in Improved MTS are as follows:

\subsubsection*{Calculation of Mahalanobis distance}
The initial step is to obtain a measurement scale which is referred to as normal observations(alternatives). Here the normal observations denote the agriculture dataset suitable for crop cultivation Table \ref{tab:table7}. The normal observations are normalized by calculating the mean and their standard deviation and the inverse of the correlation matrix of normal observations is calculated to obtain Mahalanobis Distance (MD). MD corresponding to the dataset is computed using Equation \ref{e8} \cite{22}.
\begin{equation}\mathrm{MD}=\sqrt{\frac{1}{k} Z_{i j}^{T} C^{-1} Z_{i j}}
\label{e8}
\end{equation}
Where $k$ is the number of factors, $i=1,2, ... n$ factors, $j=1,2, ... ,m$ alternatives, $Z_{ij}$ is normalized matrix calculated using the mean and standard deviation as follows:
\begin{equation}Z_{v_{j}}=\frac{X_{i j}-\bar{X}_{j}}{S_{j}}\end{equation}
where X is normal observation and factor $X_{ij}$ means $j^{th}$ characteristic of $i^{th}$ observation
$\overline{x_{i}}$ is mean value for each factor of every alternative and calculated using the formula
\begin{equation}
\overline{x_{i}}=\frac{\sum_{j=1}^{n} X_{i j}}{n}
\label{e10}
\end{equation}
${S_i}$ denotes standard deviation for each factor in normal observations and obtained using the formula
\begin{equation}S_{i}=\sqrt{\frac{\sum_{j=1}^{n}\left(X_{i j}-\overline{x_{i}}\right)^{2}}{n-1}}\end{equation}
\subsubsection*{Crop classification}
The appropriate site relevant to crop is classified using the conventional measurement scale. Every single variety of crop data (Y) is obtained from the ranking scores Table \ref{tab:table7} and made consistent by
\begin{equation}Y_{i j}=\frac{Y_{i j}-\overline{X_{j}}}{S_{j}}\end{equation}
where $\overline{X_{j}}$  is mean of column $j$ in $X$ and relative MD is computed using the formula
\begin{equation}\mathrm{MD}=\sqrt{\frac{1}{k} Y_{i j}^{T} C^{-1} Y_{i j}}\end{equation}
The Mahalanobis distance calculated for the 3 crops as per the given site dataset is presented in Table \ref{tab:table7}.
\begin{table*}[h!]
	\centering
	\caption{{Improved MTS classification results.}}
	\label{tab:table7}
	\begin{tabular}{|c|c|c|c|c|c|}
		\hline
		& \textbf{p1} & \textbf{p2} & \textbf{p3} & \textbf{p4} & \textbf{p5} \\ \hline
		$MD_p$ & \textbf{3.018445}    & \textbf{1.880574}    & \textbf{0.795746}    & \textbf{2.249553}    & \textbf{2.517626}    \\ \hline
		$MD_s$ & 47098.43    & 47355.18    & 45465.46    & 45209.61    & 47113.72    \\ \hline
		$MD_g$ & 15794.56    & 16139.26    & 2058.6      & 1848.254    & 2026.686    \\ \hline
		& \textbf{s1} & \textbf{s2} & \textbf{s3} & \textbf{s4} & \textbf{s5} \\ \hline
		$MD_p$ & 4667.18     & 4512.648    & 4556.755    & 4635.9      & 4683.685    \\ \hline
		$MD_s$ & \textbf{1.251757}    & \textbf{0.703181}    & \textbf{4.731859}    & \textbf{0.542237}    & \textbf{2.505792}    \\ \hline
		$MD_g$ & 11852.05    & 2861.864    & 11718.75    & 2983.291    & 2869.019    \\ \hline
		& \textbf{g1} & \textbf{g2} & \textbf{g3} & \textbf{g4} & \textbf{g5} \\ \hline
		$MD_p$ & 32.06872    & 5.504255    & 6.165809    & 25.13068    & 29.92342    \\ \hline
		$MD_s$ & 66.40954    & 70.91642    & 20.43124    & 19.10753    & 223.5876    \\ \hline
		$MD_g$ & \textbf{4.670055}    & \textbf{3.597042}    & \textbf{2.135998}    & \textbf{0.454961}    & \textbf{3.077487}    \\ \hline
	\end{tabular}
\end{table*}

Table \ref{tab:table7} shows the classification results of the IMTS model for 3 crops, namely paddy, sugarcane and groundnut. In Table \ref{tab:table8}, p1, p2,...p5 represents agriculture sites where paddy crop is grown, s1, s2,... s5 denotes sugarcane crop agriculture sites and g1,g2,...g3 represents agriculture dataset related to groundnut crop. Mahalanobis Distance, namely $MD_p$, $MD_s$, $MD_g$, are computed for the crops paddy, sugarcane and groundnut using the formula given in Equation \ref{e10}. Here the Mahalanobis Distance (MD) and the subscripts p,s,g are used viz. $MD_p$, $MD_s$, $MD_g$ for the crops paddy, sugarcane and groundnut respectively. The rule of least MD is the basis for classification of any agriculture site Z. Based on the MD values of the crops paddy, sugarcane and groundnut, the sites are classified. If $MD_p$ < $MD_s$ < $MD_g$, then interpretation can be made that the site dataset Z belongs to the Y type of crop.

In Table \ref{tab:table7}, for $p1, p2, …, p3$ agriculture sites, the least MD values are 3.01, 1.88, 0.79, 224, 2.51 pertaining to paddy crop. Similarly for sugarcane sites $s1, s2, ..., s3$ the least MD values 1.25, 0.70, 4.73, 0.54, 2.50 denotes the sugarcane crop. And finally the least MD values of groundnut sites $g1, g2, ..., g3$ are 4.67, 3.59, 2.13, 0.45, 3.07 shows the classified crop as groundnut. The same dataset is given to agriculture experts for classification. The results obtained from experts showed 100\% accuracy with the results obtained from the developed model. Thus the developed multiclass model is a feasible tool for classification problems.

{The multiclass models classify the three crops based on the MD calculated for each alternatives of the dataset. As well, the multiple factors considered for decision making are evaluated by considering the relative importance of each subfactor, reducing the data inconsistencies. The results obtained through the multiclass model and the agricultural experts' opinion on the dataset are similar. Since the dataset is limited, the experts were able to give their opinion were obtained certainly. Therefore, the developed multiclass model can be recommended as a feasible tool for classification problems with multiple decision criteria.}

{\subsection*{Comparative analysis of IMTS results with other classifiers}
Further, the results of IMTS is assessed by comparing with the results obtained from popular classifiers such as Naïve Bayes, Decision Table, Random Forest (Bagging with 100 iterations), AdaBoost, J48 (pruned tree with three leaves and size 5), SVM and PART. The agriculture dataset is classified using these classifiers and the proposed multiclass model under the test mode of 10 fold cross-validation\cite{purushotham2011evaluation}. The execution time for all the classifiers is not more than 0.05 seconds.} Various performance measures are used to evaluate the error rate and accuracy of selected classifiers. In order to validate the results of proposed IMTS and other classifiers, the classification accuracy, precision and recall are calculated. These metrics are calculated using the following equations:

\begin{equation}\operatorname{accuracy}=\frac{T_{P}+T_{N}}{T_{P}+T_{N}+F_{P}+F_{N}}\end{equation}
\begin{equation}\text {precision}=\frac{T_{p}}{T_{p}+F_{p}}\end{equation}
\begin{equation}\text {sensitivity}=\frac{T_{P}}{T_{P}+F_{N}}\end{equation}
$T_P$, $T_N$, $F_P$ and $F_N$ are true positive, true negative, false positive and false negative values respectively.

True positive defines test data predicted to be in decision class and is actually found in it. True negative defines test data not predicted to be in decision class and is not found in it. False-positive provides information about test data predicted to be in decision class and is not found in it. False-negative defines test data not predicted to be in decision class and is found in it. Accuracy defines the total number of correct predictions specified in percentage. Precision is defined as the total number of correct positive predictions represented in percentage. Recall defines the positive observations that are predicted as positive and specified in percentage.

\begin{figure}[h!]
	\centering
	\includegraphics[width=\linewidth]{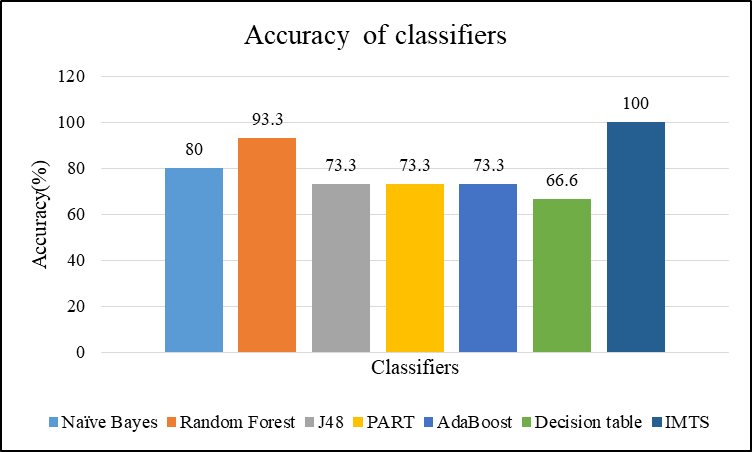}
	\caption{Accuracy scores of all classifiers.}
	\label{fig:fig3}
\end{figure}

\begin{figure}[h!]
	\centering
	\includegraphics[width=\linewidth]{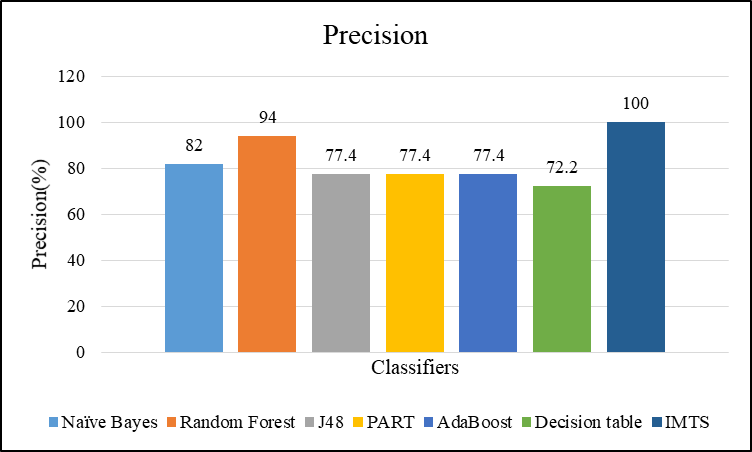}
	\caption{Precision scores of all classifiers.}
	\label{fig:fig4}
\end{figure}

\begin{figure}[h!]
	\centering
	\includegraphics[width=\linewidth]{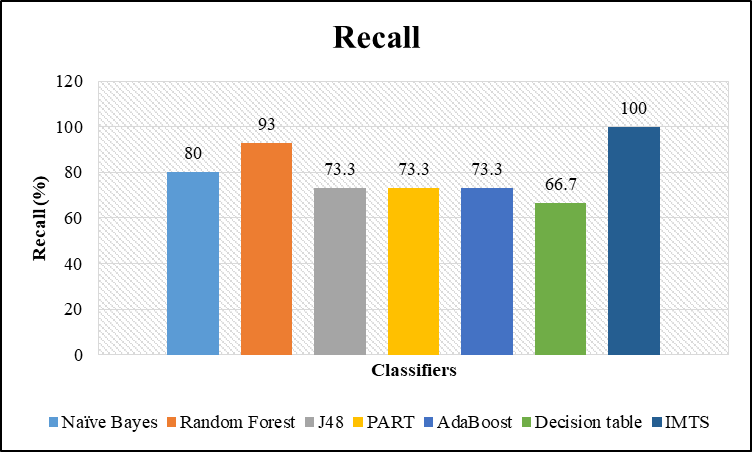}
	\caption{Recall scores of all classifiers.}
	\label{fig:fig5}
\end{figure}

The performance of the classifiers is evaluated using accuracy, precision and recall values and shown in Table \ref{tab:table8}. Accuracy, precision and recall scores of all the classifiers are represented in Figure \ref{fig:fig3} to Figure \ref{fig:fig5}.

\begin{table}[h!]
	\centering
	\caption{Evaluation metrics of various classifiers.}
	\label{tab:table8}
	\begin{tabular}{|c|c|c|c|}
		\hline
		\textbf{Classifiers} & \textbf{Accuracy} & \textbf{Precision} & \textbf{Recall} \\ \hline
		Naïve Bayes          & 80                & 82                 & 80              \\ \hline
		Random Forest        & 93.3              & 94                 & 93              \\ \hline
		J48                  & 73.3              & 77.4               & 73.3            \\ \hline
		PART                 & 73.3              & 77.4               & 73.3            \\ \hline
		AdaBoost             & 73.3              & 77.4               & 73.3            \\ \hline
		Decision Table       & 66.6              & 72.2               & 66.7            \\ \hline
		IMTS                 & 100               & 100                & 100             \\ \hline
	\end{tabular}
\end{table}

\begin{table}[h!]
	\centering
	\caption{{Evaluation of error rates of the classifiers.}}
	\label{tab:table9}
		\begin{tabular}{|c|c|c|c|c|}
			\hline
			\textbf{Classifiers}    & \textbf{MAE}   & \textbf{RMSE}  & \textbf{RAE}  & \textbf{RRSE} \\ \hline
			\textbf{Naïve Bayes}    & \textbf{19.8}  & \textbf{24.9}  & \textbf{42.9} & \textbf{50.8} \\ \hline
			\textbf{Random Forest}  & \textbf{21.15} & \textbf{25.46} & \textbf{45.7} & \textbf{51.8} \\ \hline
			\textbf{J48}            & \textbf{17.7}  & \textbf{42.1}  & \textbf{38.4} & \textbf{85.8} \\ \hline
			\textbf{PART}           & \textbf{17.7}  & \textbf{42.1}  & \textbf{38.4} & \textbf{85.8} \\ \hline
			\textbf{AdaBoost}       & \textbf{17.7}  & \textbf{42.1}  & \textbf{38.4} & \textbf{85.8} \\ \hline
			\textbf{Decision Table} & \textbf{31.3}  & \textbf{37.7}  & \textbf{67.7} & \textbf{75.5} \\ \hline
			\textbf{IMTS}           & \textbf{0}     & \textbf{0}     & \textbf{0}    & \textbf{0}    \\ \hline
		\end{tabular}%
\end{table}

\begin{figure}[h!]
	\centering
	\includegraphics[width=\linewidth]{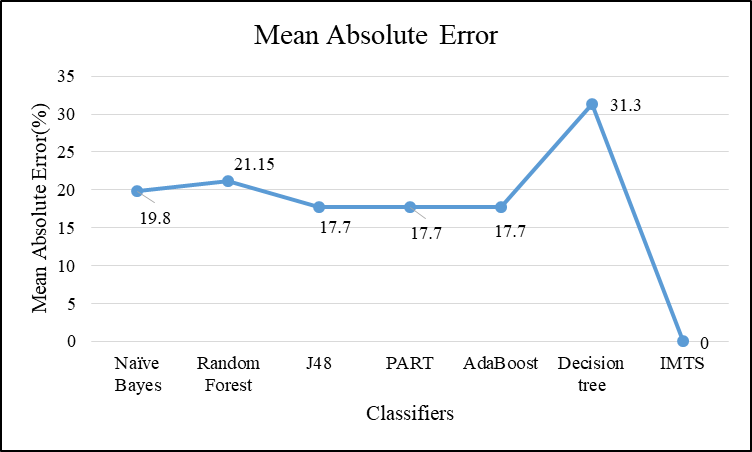}
	\caption{Mean absolute error rate of all the classifiers.}
	\label{fig:fig6}
\end{figure}

Mean Absolute Error(MAE) is used for measuring the average of the absolute difference between the set of predicted and actual values, provided each difference have identical weight. Figure \ref{fig:fig6} shows that MAE (i.e. the mean magnitude of errors) is zero for IMTS whereas Decision tree incurs 31.3\% of MAE, Random Forest has 21.15\%, and Naïve Bayes has 19.8\%. MAE for J48, PART and AdaBoost is same 17.7\%. Therefore, the multiclass model provides a 100\% match to the actual values with zero error.

\begin{figure}[h!]
	\centering
	\includegraphics[width=\linewidth]{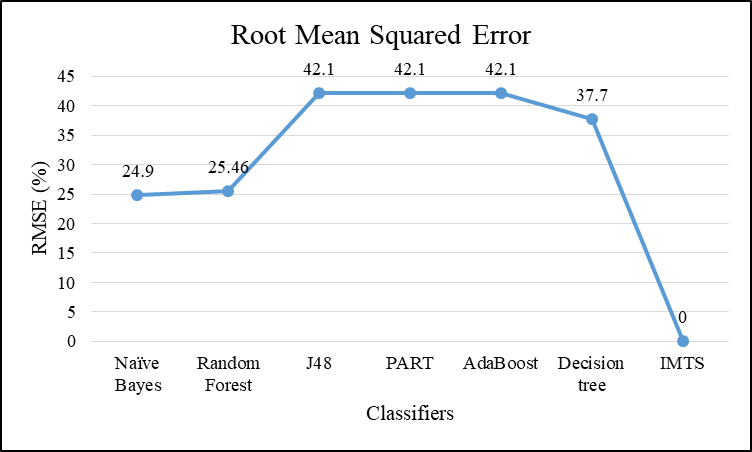}
	\caption{Root mean squared error rate of all the classifiers.}
	\label{fig:fig7}
\end{figure}

Similar to MAE, the Root Mean Square Error(RMSE) also measures the mean magnitude of the differences(i.e. errors). RMSE is the square root of the mean of the squared deviations. As RMSE is more appropriate than MAE, the proposed model has zero RMSE implying a 100\% accurate classification of crops without error whereas RMSE of remaining classifiers fall between 24\% and 42\%. RMSE graph is shown in Figure \ref{fig:fig7}. 

\begin{figure}[h!]
	\centering
	\includegraphics[width=\linewidth]{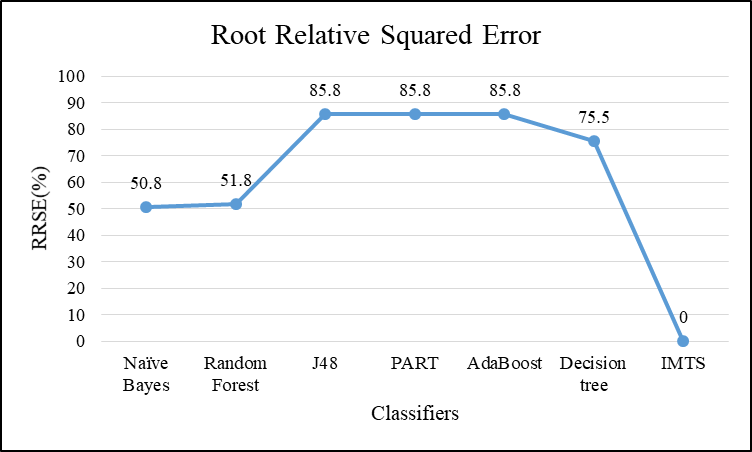}
	\caption{Root relative squared error rate of all the classifiers.}
	\label{fig:fig8}
\end{figure}

Root Relative Squared Error(RRSE) provides the squared error of the predictions that are relative to the mean of every data value. It gives accurate results than simple predictor by normalizing the values obtained from the simple predictor(Eg. Naïve or ZeroR). It divides the total squared error by dividing it with absolute squared error obtained from the simple predictor. Furthermore, by generating the square root of a normalized value, the error is reduced. The proposed model attains 0\% of RRSE as shown in Figure \ref{fig:fig8}. 

\begin{figure}[h!]
	\centering
	\includegraphics[width=\linewidth]{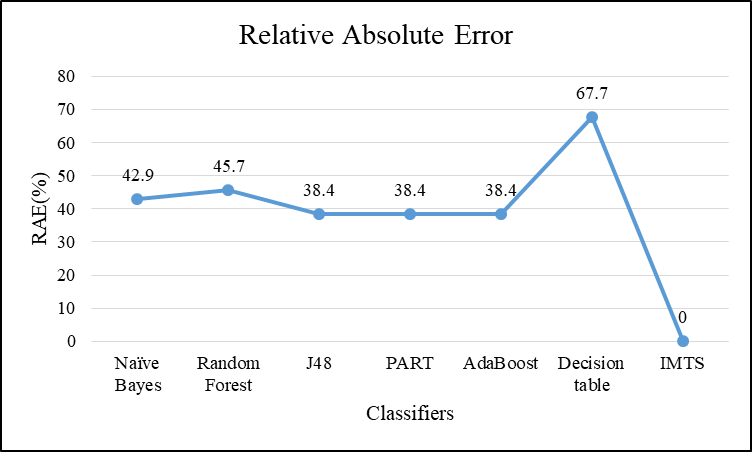}
	\caption{Relative absolute error rate of all the classifiers.}
	\label{fig:fig9}
\end{figure}

Relative Absolute Error(RAE) is similar to RRSE, which is calculated by dividing MAE by error obtained in the simple predictor. Hence, the smaller the value of RAE indicates a better prediction. Figure \ref{fig:fig8} shows that the proposed method attains 0\% of RAE, an ideal RAE value.

Mean absolute error measures the average of all the absolute errors. The root means squared error calculates the average of the magnitude of the error. Relative absolute error calculates the sum of absolute errors. The root relative squared error measures the square root of the relative squared error. The error rate of each classifier is assessed using Mean Absolute Error (MAE), Root Mean Squared Error (RMSE), Relative Absolute Error (RAE) and Root Relative Squared Error (RRSE) and shown in Table \ref{tab:table9}. The various error rates obtained for different classifiers are shown in Figure \ref{fig:fig6} to Figure \ref{fig:fig9}.
{The proposed multiclass model for classification of three different crops yields 100\% accuracy when compared to other methods. The accuracy of the classifier is assured as it considers the relative importance of each factor identified for the analysis. Grey correlation method is used for calculating the relative weights of each subfactor, and in turn the main factors are evaluated using the objective function constructed is used. This will reduce the inconsistencies in the data. The proposed model ranks the alternatives based on the least MD values. Thus by alleviating the data inconsistencies, the proposed model assures better accuracy than other models.}

\section{Conclusions}
\label{sec5}
A multiclass model is developed in this paper using the Improved Mahalanobis Taguchi System method for the classification of three crops, namely paddy, sugarcane and groundnut. Twenty-six factors are identified for the three given crops and categorized into six main factors. As the relative importance of each factor plays a major role in decision making, weights of factors are calculated using Grey correlation method. The sub-factor dataset matrix is converted to main factor data values using an objective function by applying the weights of the sub-factors. The obtained ranking score decision matrix is applied to Improved MTS for classification of three crops. Mahalanobis distance is calculated for every alternative of each crop. The least MD value forms the basic idea for the classification of agriculture site pertaining to a particular crop. The classification results of the developed model are validated by the results obtained from the agriculture experts. The multiclass model gives 100\% accuracy, recall and precision compared with other classifiers. Also, the error rates RMSE, RRSE, RAE and MAE are 0\% indicating a better prediction for the given dataset. The limitation of the model is it can be applied to decision problems with a limited number of alternatives and decision classes. {Further research can be extended by using deep neural network algorithms when high dimension dataset is applied. Feature selection methods can be applied to find a useful set of features for decision making. Other classification datasets can also be applied to test the efficiency of the developed model.}

\bibliographystyle{unsrt}
\bibliography{ref}

\EOD

\end{document}